\documentclass[runningheads]{llncs}
\usepackage[T1]{fontenc}
\usepackage{graphicx}
\usepackage{algpseudocode}
\usepackage{algorithm}
\usepackage{amsmath}
\usepackage{caption}
\usepackage{graphicx,xcolor} 
\usepackage[framemethod=tikz]{mdframed} 

\usepackage{longtable}

 \usepackage{booktabs}
 \usepackage[load-configurations=version-1]{siunitx} 

\usepackage{hyperref}
\usepackage{url}

\usepackage{booktabs}
\usepackage{multicol}
\usepackage{multirow}
\usepackage{graphicx}
\usepackage{algorithm}
\usepackage{algpseudocode}

\usepackage{tikz}
\usepackage{tikzscale}
\usepackage{xspace}
\usepackage{pgfplots}
\usepgfplotslibrary{groupplots}
\pgfplotsset{compat=1.16}
\usepackage{xparse}
\NewDocumentCommand{\Log}{o}{%
  \IfNoValueTF{#1}{}{{}^{#1}\!}\log}%

\usepackage{comment}
\usepackage{float}

\makeatletter
\def\set@curr@file#1{\def\@curr@file{#1}} 
\makeatother

\usepackage{subcaption}
\begin{document}
\title{T Cell Receptor Protein Sequences and Sparse Coding: A Novel Approach to Cancer Classification}
\titlerunning{T Cell Receptor Protein Sequences and Sparse Coding}
%
%
%

\author{Zahra Tayebi$^*$ \and
Sarwan Ali$^*$ \and
Prakash Chourasia\and
Taslim Murad \and
Murray Patterson}
\authorrunning{Z. Author et al.}


\institute{Georgia State University, Atlanta GA 30302, USA
\\
$^*$Equal Contribution}
\maketitle              
\begin{abstract}
Cancer is a complex disease marked by uncontrolled cell growth, potentially leading to tumors and metastases. Identifying cancer types is crucial for treatment decisions and patient outcomes. T cell receptors (TCRs) are vital proteins in adaptive immunity, specifically recognizing antigens and playing a pivotal role in immune responses, including against cancer. TCR diversity makes them promising for targeting cancer cells, aided by advanced sequencing revealing potent anti-cancer TCRs and TCR-based therapies.
Effectively analyzing these complex biomolecules necessitates representation and capturing their structural and functional essence. We explore sparse coding for multi-classifying TCR protein sequences with cancer categories as targets. Sparse coding, a machine learning technique, represents data with informative features, capturing intricate amino acid relationships and subtle sequence patterns.
We compute TCR sequence $k$-mers, applying sparse coding to extract key features. Domain knowledge integration improves predictive embeddings, incorporating cancer properties like Human leukocyte antigen (HLA) types, gene mutations, clinical traits, immunological features, and epigenetic changes. Our embedding method, applied to a TCR benchmark dataset, significantly outperforms baselines, achieving 99.8\% accuracy.
Our study underscores sparse coding's potential in dissecting TCR protein sequences in cancer research.

\keywords{Cancer classification  \and TCR sequence\and embeddings.}
\end{abstract}
\section{Introduction}
\begin{figure}[h!]
    \centering
  \includegraphics[scale=0.45]{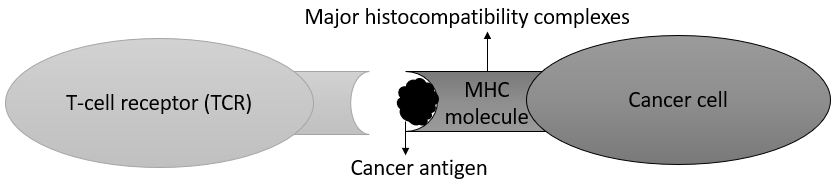}
  \caption{T-cells mount a targeted immune response against the invading pathogen of cancerous cells.}
  \label{fig_TCR_MHC}
\end{figure}

T cell receptors (TCRs) play a crucial role in the immune response by recognizing and binding to antigens presented by major histocompatibility complexes (MHCs) on the surface of infected or cancerous cells, Figure~\ref{fig_TCR_MHC}\cite{shah2021t}. The specificity of TCRs for antigens is determined by the sequence of amino acids that make up the receptor, which is generated through a process of genetic recombination and somatic mutation. This enables T cells to produce a diverse repertoire of receptors capable of recognizing a wide range of antigens~\cite{courtney2018tcr}.

MHC molecules bind to peptide fragments that originate from pathogens and present them on the cell surface. This allows for recognition by the corresponding T cells~\cite{janeway2001major}. TCR sequencing involves analyzing the DNA or RNA sequences that code for the TCR protein on T cells, and it can be used to identify changes in the TCR repertoire that occur in response to cancer, as well as to identify specific TCR sequences that are associated with particular types of cancer~\cite{kidman2020characteristics}.

Embedding-based methods, exemplified by Yang et al.\cite{yang2020prediction}, transform protein sequences into low-dimensional vector representations. They find applications in classification and clustering\cite{chourasia2023reads2vec}. These embeddings empower classifiers to predict functional classes for unlabeled sequences~\cite{iqbal2014efficient} and identify related sequences with shared features~\cite{yang2020prediction}.


Embedding-based methods in protein sequence analysis face challenges related to generalizability and complexity~\cite{mikolov2013efficient}. Deep learning techniques, including Convolutional Neural Networks (CNNs), Recurrent Neural Networks (RNNs), and transformer networks, offer promising solutions~\cite{nambiar2020transforming}. These models, trained on extensive and diverse datasets, enhance generalization and enable accurate predictions for new sequences. However, factors like data quality, model architecture, and optimization strategies significantly influence performance~\cite{min2021pre}. Furthermore, the interpretability of deep learning models remains a challenge, necessitating careful evaluation. While deep learning enhances generalizability, it may not completely resolve the problem.

In our study, we propose a multi-class classification approach for predicting cancer categories from TCR protein sequences. We employ sparse coding~\cite{olshausen2004sparse}, a machine-learning technique using a dictionary of sparse basis functions. By representing TCR protein sequences as sparse linear combinations of these basis functions, we capture inherent data structure and variation, improving cancer classification accuracy. Preprocessing involves encoding amino acid sequences as numerical vectors and utilizing a $k$-mer dictionary for sparse coding.

We assess our approach using multiple metrics and benchmark it against state-of-the-art methods. Our results clearly show its superior classification performance and highlight its key properties: invariance, noise robustness, interpretability, transferability, and flexibility.

In general, our contributions to this paper are the following:
\begin{enumerate}
    \item We introduce an efficient method using sparse coding and $k$-mers to create an alignment-free, fixed-length numerical representation for T cell sequences. Unlike traditional one-hot encoding, we initially extract $k$-mers and then apply one-hot encoding, preserving context through a sliding window approach.
    \item Leveraging domain knowledge, we enhance supervised analysis by combining it with sparse coding-based representations. This fusion enriches the embeddings with domain-specific information, preserving sequence details, facilitating feature selection, promoting interpretability, and potentially boosting classification accuracy.
    \item Our proposed embedding method achieves near-perfect predictive performance, significantly surpassing all baselines in classification accuracy. This highlights the effectiveness of our approach in distinguishing T cell sequences despite their short length.
    
\end{enumerate}

Considering this, we will proceed as follows: Section~\ref{sec_related_work} covers related work, Section~\ref{sec_proposed_approach} presents our approach, Section~\ref{sec_experimental_setup} details the dataset and experimental setup, Section~\ref{sec_results} reports results, and Section~\ref{sec_conclusion} concludes the paper.

\section{Related Work}\label{sec_related_work}
TCR (T-cell receptor) sequencing data has emerged as a critical asset in biomedical research, revolutionizing our understanding of the immune system and its applications in fields such as immunotherapy and disease response. This technology has enabled the identification of neoantigens, which are crucial targets for immunotherapy in cancer treatment~\cite{van2020tumor}. Moreover, it has played a pivotal role in decoding the adaptive immune response to viral diseases like the formidable SARS-CoV-2~\cite{gittelman2022longitudinal}. TCR sequencing has also contributed significantly to the identification of T-cell clones associated with tumor regression in response to immunotherapy~\cite{lu2019single}.

In the realm of biological data analysis, classification methods have proven invaluable for deciphering protein sequences and their implications in cancer research. Techniques like Support Vector Machines (SVM), Random Forest, and Logistic Regression have been instrumental in grouping protein sequences with shared features, aiding in the discovery of subpopulations related to specific cancer types~\cite{hoadley2018cell}. For instance, SVM was employed to classify TCR sequences for predicting disease-free and overall survival in breast cancer patients~\cite{bai2018use}. K-means classification has demonstrated promise in distinguishing colorectal cancer patients from healthy individuals based on protein sequence analysis~\cite{bae2021feature}.

Deep learning methods have ushered in a new era of protein property prediction, encompassing structure, function, stability, and interactions~\cite{wan2019deepcpi}. Convolutional neural networks (CNNs) have proven effective in forecasting protein structure and function from amino acid sequences~\cite{bileschi2019using}. Recurrent neural networks (RNNs) excel at modeling temporal dependencies and long-range interactions within protein sequences~\cite{zhang2021recurrent}. The Universal Representation of Amino Acid Sequences (UniRep) employs an RNN architecture to embed protein sequences, yielding promising results across various biological data analysis tasks~\cite{alley2019unified}. Innovative methods like ProtVec use word embedding techniques to create distributed representations of protein sequences for function prediction~\cite{ostrovsky2021immune2vec}. SeqVec introduces an alternative by embedding protein sequences using a hierarchical representation that captures both local and global features~\cite{heinzinger2019modeling}.

Lastly, ESM (Evolutionary Scale Modeling) employs a transformer-based architecture to encode protein sequences, forming part of the ESM family of protein models~\cite{lin2023evolutionary}. It is trained on extensive data encompassing protein sequences and structures to predict the 3D structure of a given protein sequence~\cite{hu2022exploring}. While deep learning enriches the generalizability of embedding-based techniques for protein sequence analysis, robust evaluation and consideration are essential to ensure the reliability of outcomes~\cite{lin2023evolutionary}. 

\section{Proposed Approach}\label{sec_proposed_approach}
In this section, we delve into the utilization of cancer-related knowledge, introduce our algorithm, and offer an overview through a flowchart.
\subsection{Incorporating Domain Knowledge}
This section gives some examples of the additional property values for the four cancers we mentioned.
Many factors can increase the risk of developing cancer, including Human leukocyte antigen (HLA) types, gene mutations, clinical characteristics, immunological features, and epigenetic modifications.

\textbf{HLA types: } HLA genes encode cell surface proteins presenting antigens to T cells. Mutations in these genes due to cancer can hinder antigen presentation, enabling unchecked cancer growth~\cite{schaafsma2021pan}.

\textbf{Gene mutations: }
Besides HLA, mutations in genes like TP53 and BRCA1/2 raise cancer risks (breast, ovarian, colorectal) too~\cite{peshkin2011brca1}. 

\textbf{Clinical characteristics: }
Age, gender, and family history influence cancer risk~\cite{lee2019boadicea}. Certain cancers are age-specific, some gender-biased, while family history also shapes susceptibility.

\textbf{Immunological features: }
The immune system crucially impacts cancer outcomes. Immune cells recognize and eliminate cancer cells~\cite{de2006paradoxical}. Yet, cancer cells might evade immune surveillance, growing unchecked. Immune cell presence and activity within tumors influence cancer progression and treatment response~\cite{gonzalez2018roles}.

\textbf{Epigenetic modifications: }
Epigenetic changes, like DNA methylation, contribute to cancer development and progression. DNA methylation, adding a methyl group to DNA's cytosine base (CpG sites), leads to gene silencing or altered expression~\cite{kelly2010epigenetic}.

\subsection{Cancer Types and Immunogenetics}
Cancer types exhibit intricate connections between immunogenetics, clinical attributes, and genetic factors. Breast cancer, a prevalent malignancy, demonstrates HLA associations such as HLA-A2, HLA-B7, and HLA-DRB1*15:01 alleles, alongside gene mutations (BRCA1, BRCA2, TP53, PIK3CA) that elevate risk~\cite{liang2021relationships,johnson2007counting}. Clinical attributes encompass tumor size, grade, hormone receptor presence, and HER2 status. Tumor-infiltrating lymphocytes (TILs) and immune checkpoint molecules (PD-1, PD-L1, CTLA-4) play a substantial role in influencing breast cancer outcomes~\cite{loibl2017her2,stanton2016clinical}. Similarly, colorectal cancer (CRC), originating in the colon or rectum, shows associations with HLA alleles (HLA-A11, HLA-B44) and gene mutations (APC, KRAS, TP53, BRAF) that impact its development~\cite{dunne2020characterising,fodde2002apc}. The intricacies of CRC involve tumor attributes, lymph node involvement, TILs, and immune checkpoint molecules, all influencing disease progression~\cite{rotte2019combination}. Liver cancer (hepatocellular carcinoma), primarily affecting liver cells, displays connections with HLA alleles (HLA-A2, HLA-B35) and gene mutations (TP53, CTNNB1, AXIN1, ARID1A)\cite{makuuchi1993surgery,bufe2022pd}. Epigenetic DNA methylation abnormalities in CDKN2A, MGMT, and GSTP1 genes contribute to liver cancer's development~\cite{zhu2005altered}. Urothelial cancer, affecting bladder, ureters, and renal pelvis cells, is associated with HLA types (HLA-A2, HLA-B7) and gene mutations (FGFR3, TP53, RB1, PIK3CA)\cite{carosella2015systematic}.

\subsection{Algorithmic Pseudocode and Flowchart}





   





\begin{figure}[h!]
\centering
  \includegraphics[scale = 0.35] {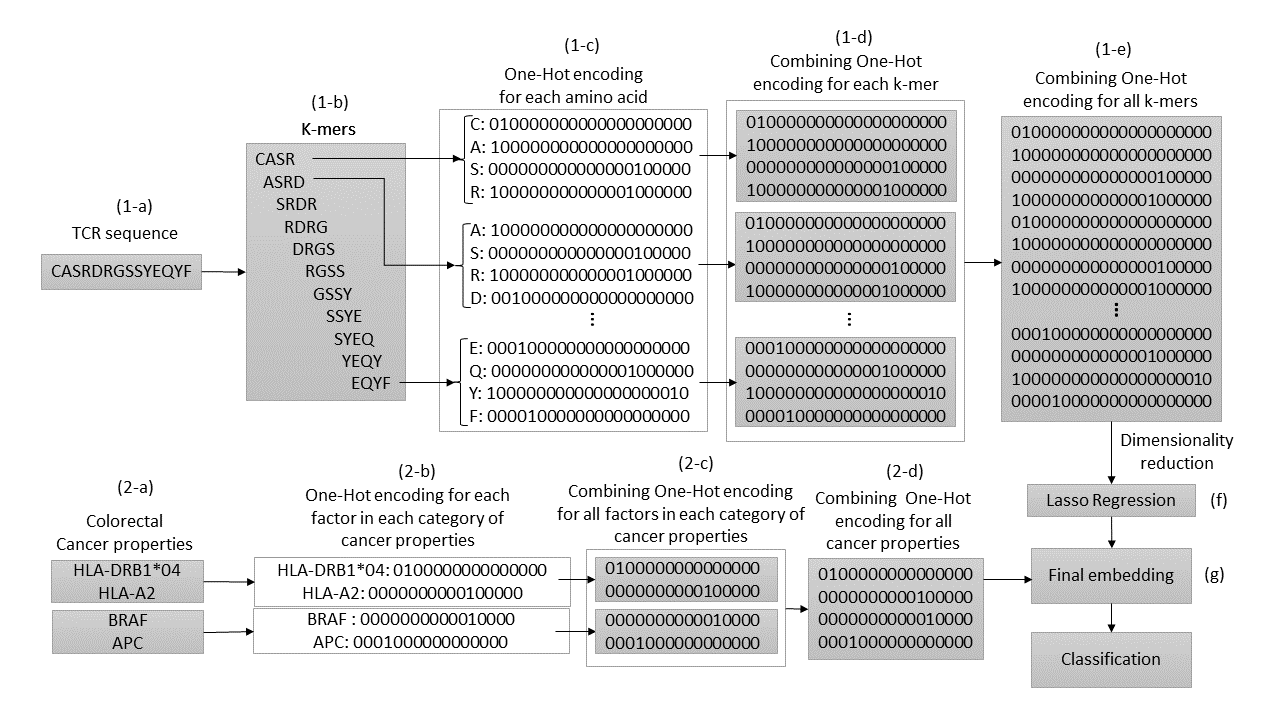}
  \caption{Flowchart of TCR sequence analysis}
  \label{fig_TCR}
\end{figure}

The pseudocode to compute sparse coding + $k$-mers-based embedding is given in Algorithm~\ref{algo_sparse_coding}. This algorithm takes a set of sequences $S = \{s_1, s_2, \ldots, N \}$ and $k$, where $N$ is the number of sequences and $k$ is the length of $k$-mers. The algorithm computed the sparse embedding by iterating over all sequences and computing the set of $k$-mers for each sequence; see Figure~\ref{fig_TCR} (1-b). Then it iterates over all the $k$-mers, computes the one-hot encoding (OHE) based representation for each amino acid within a $k$-mer (Figure~\ref{fig_TCR} (1-c)), and concatenates it with the OHE embeddings of other amino acids within the $k$-mer (Figure~\ref{fig_TCR} (1-d)). Finally, the OHE embeddings for all $k$-mers within a sequence are concatenated to get the final sparse coding-based representation (Figure~\ref{fig_TCR} (1-e)). To avoid the curse of dimensionality, we used Lasso regression as a dimensionality reduction technique~\cite{ranstam2018lasso}, (Figure~\ref{fig_TCR} (f)). The objective function, which we used in lasso regression, is the following: \text{min(Sum of square residuals} + $\alpha \times \vert slope \vert$)

In the above equation, the $\alpha \times \vert slope \vert$ is referred to as the penalty terms, which reduces the slope of insignificant features to zero.
We use $k=4$ for experiments, which is decided using the standard validation set approach.

\begin{algorithm}[h!]


\begin{algorithmic}[1]
\scriptsize
\Statex \textbf{Input: } set of sequences S, $k$-mers length k
\Statex \textbf{Output: } SparseEmbedding

\State totValues = $21$ 
\Comment{unique amino acid characters}

\State final\_sparse\_embedding $\gets$ []

\For{i $\gets$ 0 to $\vert S \vert$}

\State seq $\gets$ S[i]

\State kmers $\gets$ \Call{generateKmers}{seq, k} \Comment{generate set of $k$-mers}

\State encoded\_kmers $\gets$ []

\For{kmer \textbf{in} kmers}

    \State encodedVec $\gets$ np.zeros(totValues$^{k}$) \Comment{$21^3 = 9261$ dimensional vector}

    \For{i, aa \textbf{in} enumerate(kmer)}

        \State pos $\gets$ i $\times$ totValues$^{k-i-1}$

        \State encodedVec[pos:pos+totValues] $\gets$ \Call{OneHotEncoding}{aa}

    \EndFor

\State encoded\_kmers.append(encodedVec)

\EndFor
\State final\_sparse\_embedding.append(np.array(encoded\_kmers).flatten())

\EndFor

\State SparseEmbedding $\gets$ \Call{LassoRegression}{final\_sparse\_embedding} 
\Comment{dim. reduction}

\State \textbf{return} SparseEmbedding
\end{algorithmic}
\caption{Sparse Coding Algorithm}
\label{algo_sparse_coding}
\end{algorithm}

The same process repeated while we considered domain knowledge about cancer properties like HLA types, gene mutations, clinical characteristics, immunological
features, and epigenetic modifications. For each property, we generate a one-hot encoding-based representation, where the length of the vector equals the total number of possible property values, see Figure~\ref{fig_TCR} (2-b). All OHE representations of the property values are concatenated in the end to get final representations for all properties we consider from domain knowledge; see Figure~\ref{fig_TCR} (2-c) to (2-d). The resultant embeddings from Sparse coding (from Algorithm~\ref{algo_sparse_coding}) and domain knowledge are concatenated to get the final embedding. These embeddings are used to train machine-learning models for supervised analysis (Figure~\ref{fig_TCR} (g))

\section{Experimental Setup}\label{sec_experimental_setup}
In this section, we describe the details related to the dataset,  baseline models, evaluation metrics, and data visualization. We employ a variety of machine learning classifiers, encompassing Support Vector Machine (SVM), Naive Bayes (NB), Multi-Layer Perceptron (MLP), K-Nearest Neighbors (KNN) using K=3, Random Forest (RF), Logistic Regression (LR), and Decision Tree (DT). We used a 70-30\%
train-test split based on stratified sampling to do classification. From the training
set, we use 10\% data as a validation set for hyperparameter tuning. Performance evaluation of the classifiers employed various
evaluation metrics. 
The results are computed 5 times, and the average results are reported. The experiments were performed on a system with an Intel(R) Core i5 processor @ $2.10$ GHz and a Windows 10 $64$ bit operating system with 32 GB of memory. The model is implemented using Python, which is available online~\footnote{\url{https://github.com/zara77/T-Cell-Receptor-Protein-Sequences-and-Sparse-Coding.git}}.

\subsection{Dataset Statistics}
We source sequence data from TCRdb, a comprehensive $T$-cell receptor sequence database with robust search functionality~\cite{chen2021tcrdb}. TCRdb consolidates 130+ projects and 8000+ samples, comprising an extensive sequence repository. Our focus identifies four prevalent cancer types by incidence rates. Extracting 23331 TCR sequences, we maintain the original data proportion via Stratified ShuffleSplit. Table~\ref{tbl_data_statistics} offers dataset statistics.
Our embedding approach employs TCR sequences, cancer names, and cancer properties. Table~\ref{tbl_gene_mutations} exemplifies TCR sequences, cancer names, and gene mutations specific to each cancer type. This includes $4$ cancers with respective gene mutations that impact risk. 

\begin{table}[h!]
    \centering
    \begin{subtable}{0.5\textwidth}
        \centering
        \resizebox{0.8\linewidth}{!}{%
            \begin{tabular}{ccccc}
            \toprule
            & & \multicolumn{3}{c}{Sequence Length Statistics} \\
            \cmidrule{3-5}
            Cancer Name & Number of Sequences & Min. & Max. & Average \\
            \midrule \midrule
            Breast & 4363 & 8 & 20 & 14.2264 \\
            Colorectal &  10947 & 7 & 26 & 14.5573 \\
            Liver &  3520 & 8 & 20 & 14.3005 \\
            Urothelial &  4501 & 7 & 24 & 14.6538 \\
            \midrule
            Total & 23331 & - & - & - \\
            \bottomrule
            \end{tabular}
        }
        \caption{}
        \label{tbl_data_statistics}
    \end{subtable}%
    \begin{subtable}{0.5\textwidth}
        \centering
        \resizebox{0.8\linewidth}{!}{%
            \begin{tabular}{ccc}
            \toprule
            Sequence & Cancer Name & Gene Mutation \\
            \midrule \midrule
            CASSRGQYEQYF    & Breast      & BRCA1, BRCA2, TP53, PIK3CA  \\
            CASSLEAGRAYEQYF & Colorectal  & APC, KRAS, TP53, BRAF \\
            CASSLGSGQETQYF  & Liver & ~TP53, CTNNB1, AXIN1  \\
            CASSGQGSSNSPLHF & Urothelial  & FGFR3,TP53, RB1, PIK3CA \\
            \bottomrule
            \end{tabular}
        }
        \caption{}
        \label{tbl_gene_mutations}
    \end{subtable}
    \caption{(a) Dataset Statistics. (b) Examples of sequences for different cancer types, including Breast, Colorectal, Liver, and Urothelial, along with their respective gene mutations.}
    \label{tbl_combined}
\end{table}

To assess the effectiveness of our proposed system, we employed a variety of baseline methods, categorized as follows:
(i) Feature Engineering methods: (One Hot Encoding (OHE)~\cite{kuzmin2020machine}, Spike2Vec~\cite{ali2021spike2vec}, PWM2Vec~\cite{ali2022pwm2vec}, and Spaced $k$-mers~\cite{singh2017gakco}), (ii) Kernel
method (String kernel~\cite{farhan2017efficient}), (iii) Neural network (Wasserstein Distance Guided Representation Learning (WDGRL)~\cite{shen2018wasserstein} and AutoEncoder~\cite{xie2016unsupervised}), (iv) Pre-trained Larger Language Model (SeqVec~\cite{heinzinger2019modeling}),
and (v) Pre-trained Transformer (ProteinBert~\cite{BrandesProteinBERT2022}).

\subsection{Data Visualization}
\begin{figure}[h!]
 \centering
    \begin{subfigure}{.16\textwidth}
        \centering
        \includegraphics[scale=0.035]{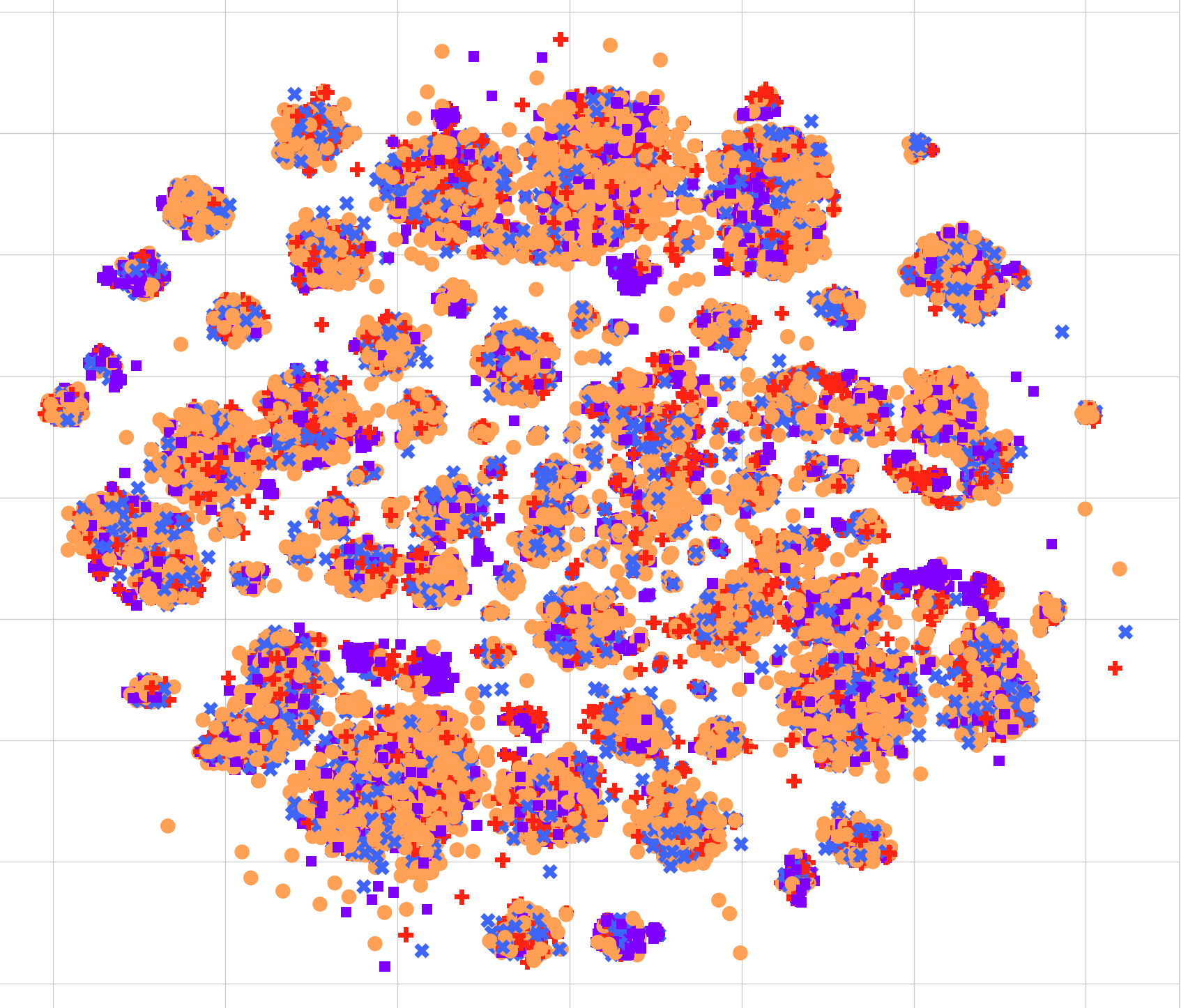}
        \caption{OHE}
        \label{}
    \end{subfigure}%
    \begin{subfigure}{.17\textwidth}
        \centering
        \includegraphics[scale=0.035]{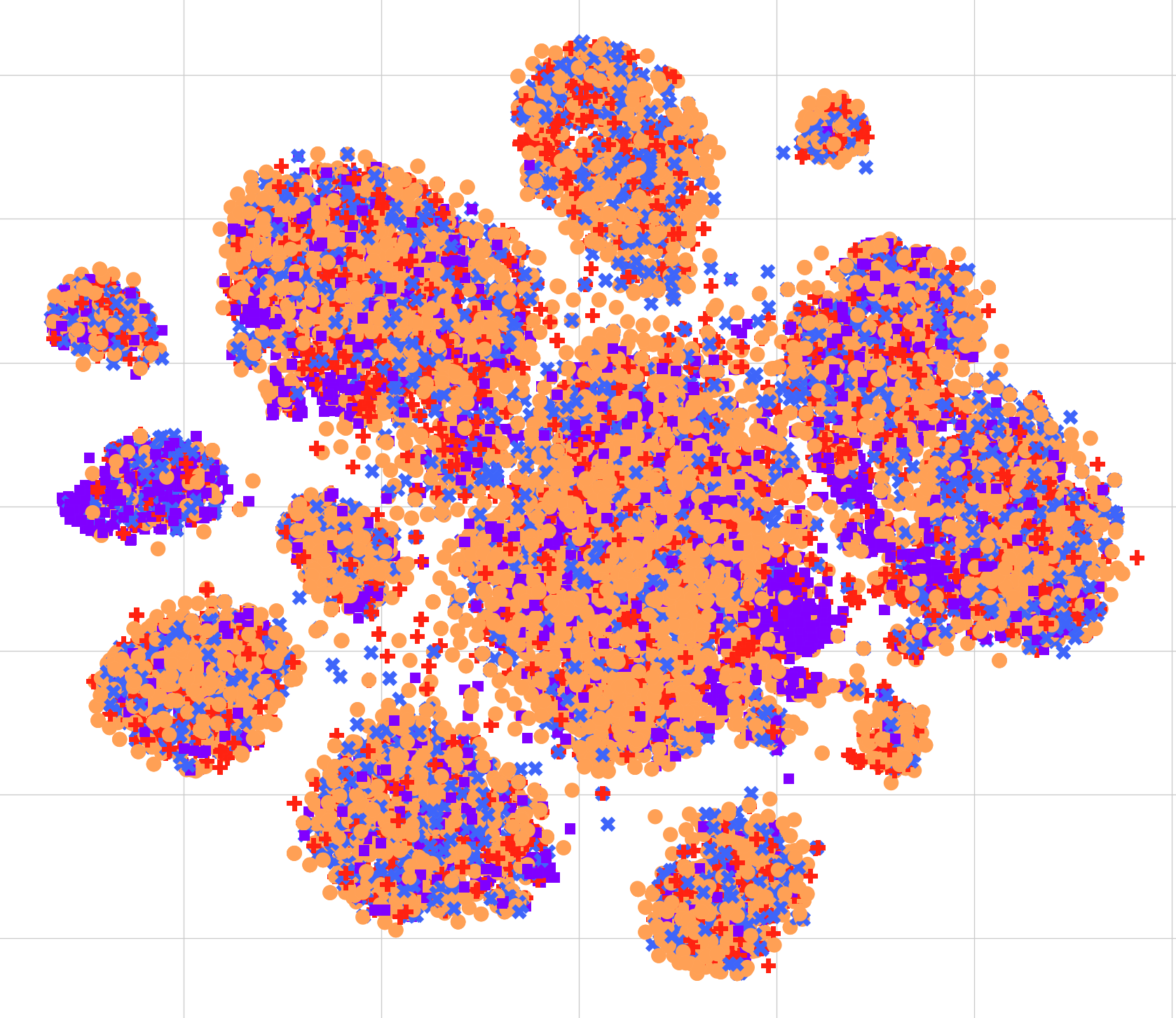}
        \caption{Spike}
        \label{}
    \end{subfigure}%
    \begin{subfigure}{.16\textwidth}
        \centering
        \includegraphics[scale=0.035]{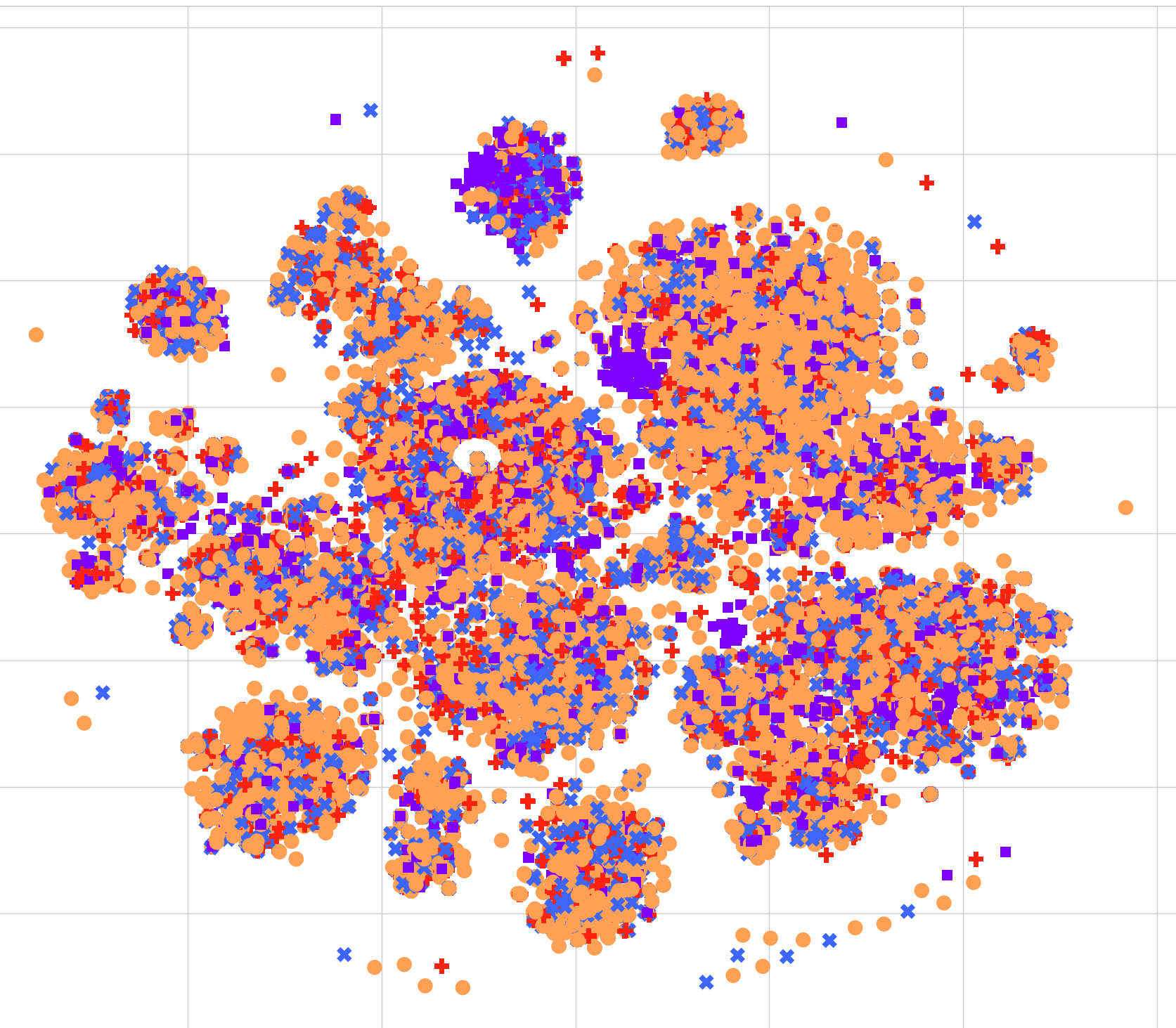}
        \caption{PWM}
        \label{}
    \end{subfigure}%
    \begin{subfigure}{.17\textwidth}
        \centering
        \includegraphics[scale=0.035]{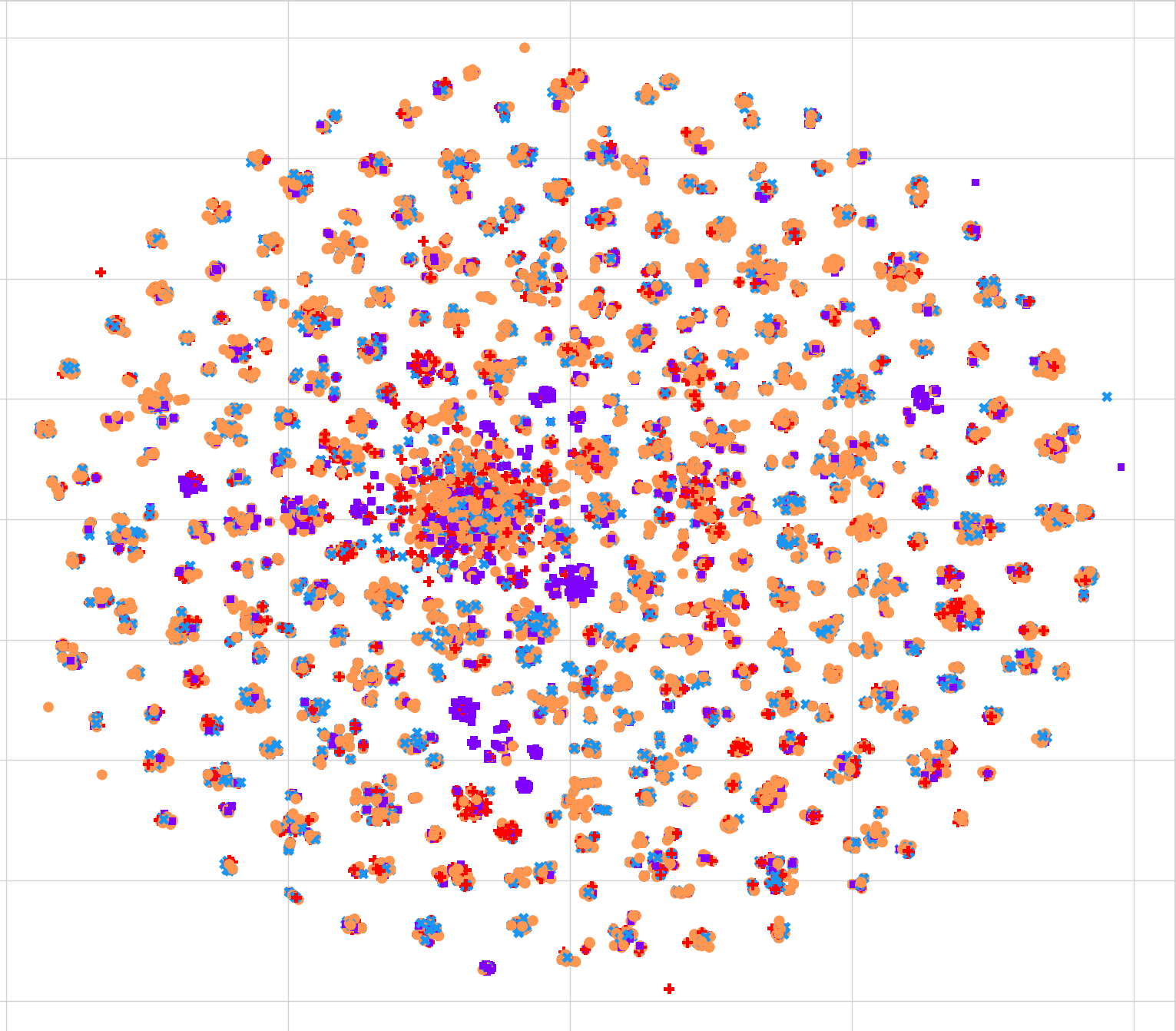}
        \caption{S $k$-mers}\label{}
    \end{subfigure}%
    \begin{subfigure}{.17\textwidth}
        \centering
        \includegraphics[scale=0.035]{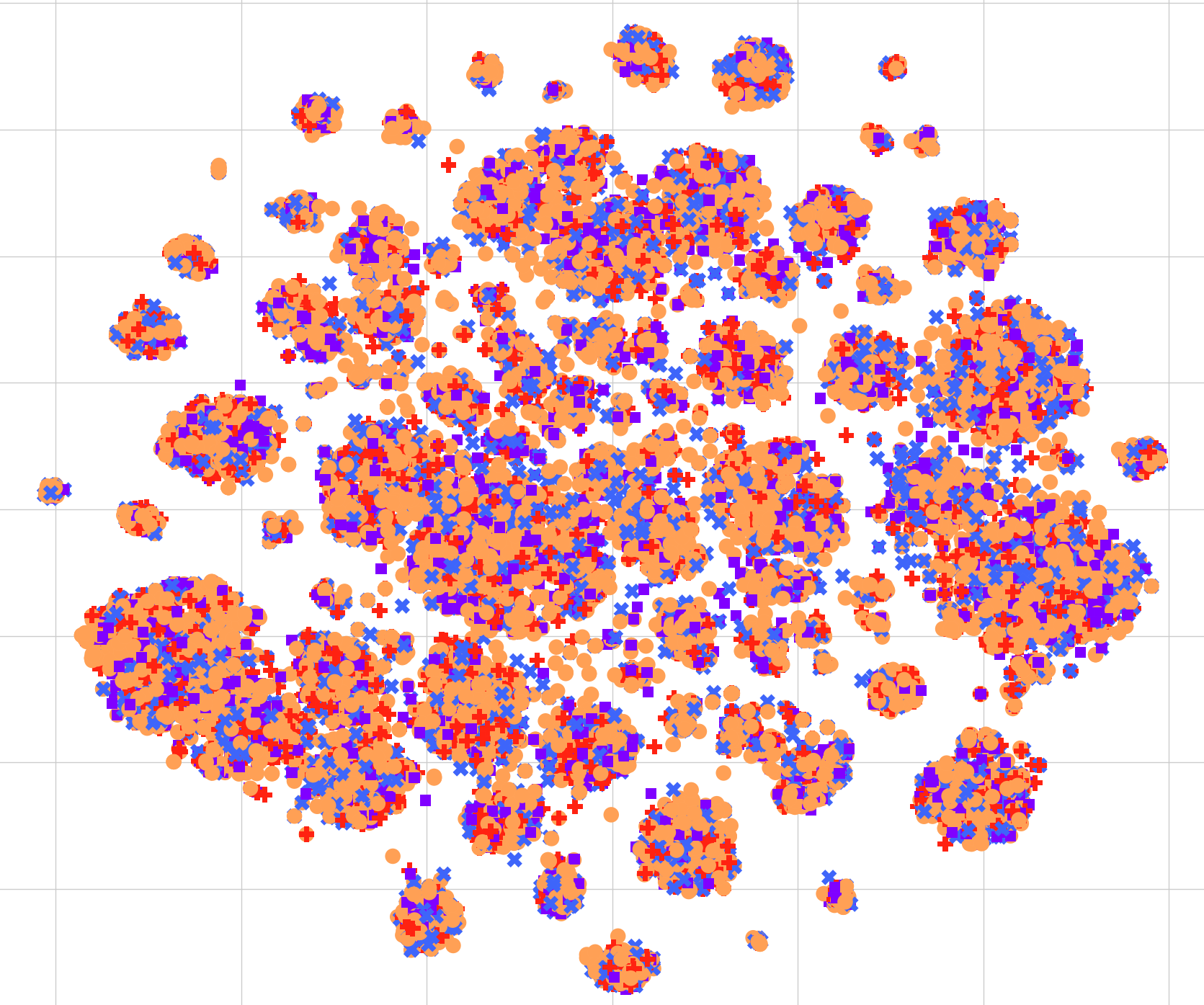}
        \caption{AE}
        \label{}
    \end{subfigure}%
    \begin{subfigure}{.17\textwidth}
        \centering
        \includegraphics[scale=0.035]{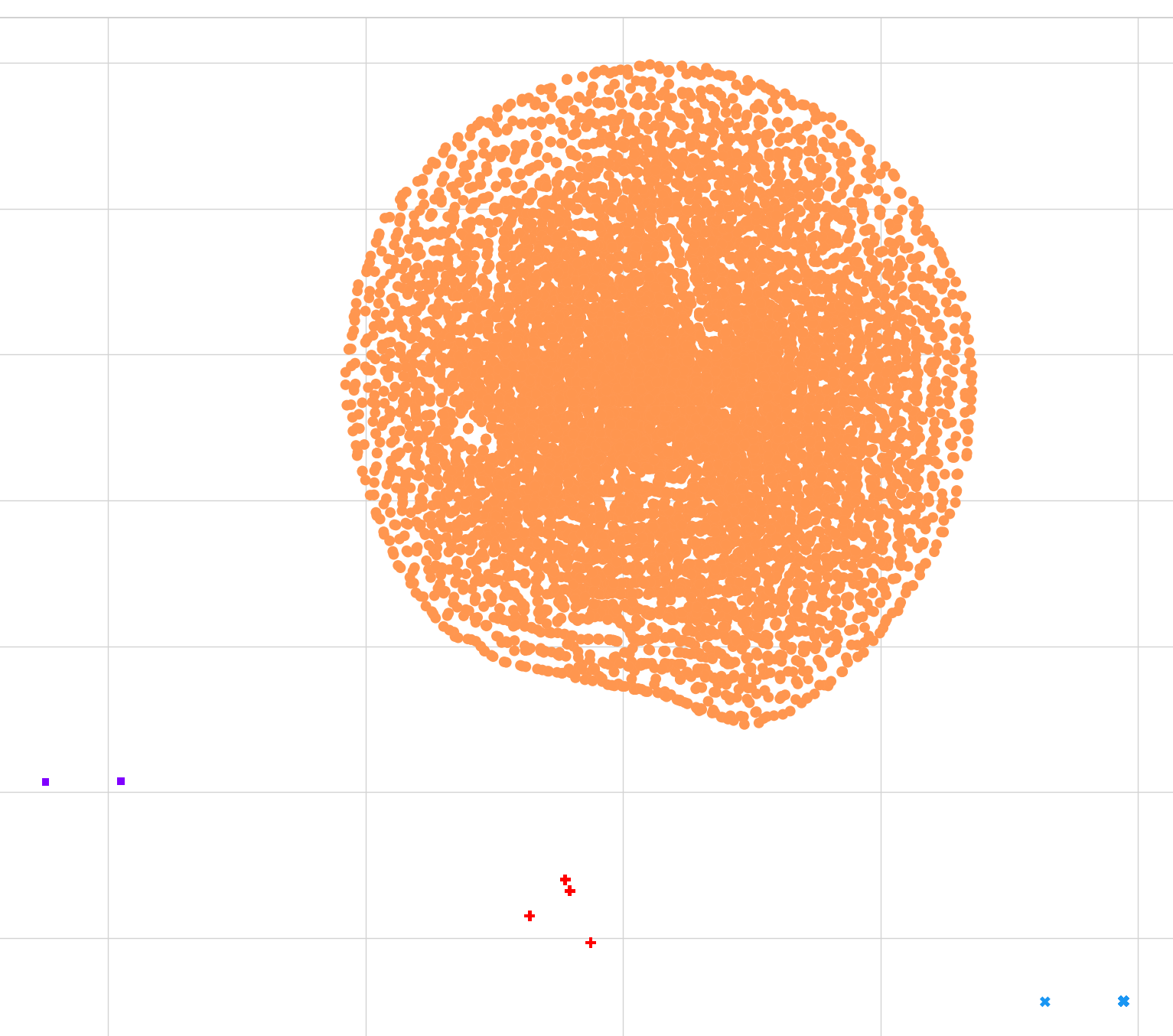}
        \caption{SC}\label{}
    \end{subfigure}%
    \\
    \begin{subfigure}{1\textwidth}
        \centering
        \includegraphics[scale=0.3]{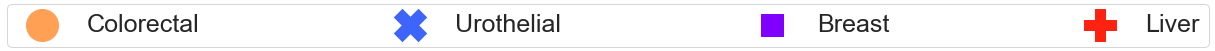}
        \caption*{}
        \label{}
    \end{subfigure}%
    \caption{t-SNE plots for different feature embedding methods. The figure is best seen in color. Subfigures (b), (c), (d), (e), and (f) are for Spike2Vec, PWM2Vec, Spaced $k$-mers, Autoencoder (AE), and Sparse Coding (SC), respectively.}
    \label{tsne_plots}
\end{figure}

To see if there is any natural (hidden) clustering in the data, we use t-distributed stochastic neighbor embedding (t-SNE)~\cite{van2008visualizing}, which maps input sequences to 2D representation. 
The t-SNE plots for different embedding methods are shown in Figure~\ref{tsne_plots} for One hot encoding (OHE), Spike2Vec, PWM2Vec, Spaced $K$-mer, Autoencoder, and Sparse Coding, respectively. We can generally observe that OHE, Spike2Vec, PWM2Vec, and Autoencoder show smaller groups for different cancer types. However, the Spaced $k$-mers show a very scattered representation of data. Moreover, the proposed Sparse Coding approach does not show any scattered representation, preserving the overall structure better.

\begin{table}[h!]
    \centering
    \resizebox{0.62\textwidth}{!}{
    \begin{tabular}{p{1.8cm}lp{1.1cm}p{1.1cm}p{1.1cm}p{1.9cm}p{1.9cm}p{1.9cm}|p{1.6cm}}
    \toprule
        \multirow{2}{*}{Embeddings} & \multirow{2}{*}{Algo.} & \multirow{2}{*}{Acc. $\uparrow$} & \multirow{2}{*}{Prec. $\uparrow$} & \multirow{2}{*}{Recall $\uparrow$} & \multirow{2}{1.7cm}{F1 (Weig.) $\uparrow$} & \multirow{2}{1.9cm}{F1 (Macro) $\uparrow$} & \multirow{2}{1.2cm}{ROC AUC $\uparrow$} & Train Time (sec.) $\downarrow$ \\
        \midrule \midrule
        \multirow{7}{1.2cm}{OHE}
         & SVM & 0.5101 & 0.5224 & 0.5101 & 0.4073 & 0.3152 & 0.5592 & 790.5622 \\
 & NB & 0.2036 & 0.3533 & 0.2036 & 0.0917 & 0.1213 & 0.5107 & 1.0560 \\
 & MLP & 0.4651 & 0.4368 & 0.4651 & 0.4370 & 0.3714 & 0.5764 & 221.0638 \\
 & KNN & 0.4464 & 0.4044 & 0.4464 & 0.4100 & 0.3354 & 0.5617 & 7.2748 \\
 & RF & 0.5156 & 0.5003 & 0.5156 & 0.4521 & 0.3751 & 0.5824 & 18.7857 \\
 & LR & 0.5143 & 0.5241 & 0.5143 & 0.4327 & 0.3492 & 0.5701 & 61.8512 \\
 & DT & 0.4199 & 0.4129 & 0.4199 & 0.4160 & 0.3616 & 0.5737 & 1.0607 \\
        
        \cmidrule{2-9}
        \multirow{7}{1.2cm}{Spike2Vec}
         & SVM & 0.4309 & 0.4105 & 0.4309 & 0.4157 & 0.3543 & 0.5711 & 19241.67 \\
 & NB & 0.2174 & 0.3535 & 0.2174 & 0.1931 & 0.2081 & 0.5221 & 6.1309 \\
 & MLP & 0.4191 & 0.4081 & 0.4191 & 0.4128 & 0.3490 & 0.5671 & 227.2146 \\
 & KNN & 0.4397 & 0.4105 & 0.4397 & 0.4087 & 0.3400 & 0.5673 & 40.4765 \\
 & RF & 0.5183 & 0.5078 & 0.5183 & 0.4519 & 0.3738 & 0.5836 & 138.6850 \\
 & LR & 0.4404 & 0.4189 & 0.4404 & 0.4251 & 0.3626 & 0.5728 & 914.7739 \\
 & DT & 0.4510 & 0.4307 & 0.4510 & 0.4365 & 0.3745 & 0.5805 & 40.3568 \\
        
        \cmidrule{2-9}
        \multirow{7}{1.2cm}{PWM2Vec}
         & SVM & 0.4041 & 0.3760 & 0.4041 & 0.3836 & 0.3138 & 0.5452 & 10927.61 \\
 & NB & 0.2287 & 0.3758 & 0.2287 & 0.2109 & 0.2220 & 0.5273 & 23.9892 \\
 & MLP & 0.4053 & 0.3820 & 0.4053 & 0.3888 & 0.3226 & 0.5467 & 253.6387 \\
 & KNN & 0.4454 & 0.3890 & 0.4454 & 0.3930 & 0.3069 & 0.5475 & 10.2005 \\
 & RF & 0.4994 & 0.4745 & 0.4994 & 0.4370 & 0.3548 & 0.5716 & 126.5780 \\
 & LR & 0.4143 & 0.3861 & 0.4143 & 0.3937 & 0.3237 & 0.5484 & 991.8051 \\
 & DT & 0.4339 & 0.4078 & 0.4339 & 0.4140 & 0.3496 & 0.5636 & 34.9344 \\

        \cmidrule{2-9}
        \multirow{7}{1.2cm}{Spaced $k$-mers}
         & SVM & 0.5109 & 0.5221 & 0.5109 & 0.4095 & 0.3143 & 0.5582 & 824.215 \\
 & NB & 0.2157 & 0.3713 & 0.2157 & 0.1296 & 0.1510 & 0.5144 & 0.1883 \\
 & MLP & 0.4524 & 0.4203 & 0.4524 & 0.4236 & 0.3550 & 0.5663 & 207.685 \\
 & KNN & 0.4527 & 0.4078 & 0.4527 & 0.4132 & 0.3351 & 0.5607 & 3.3905 \\
 & RF & 0.5204 & 0.5233 & 0.5204 & 0.4294 & 0.3393 & 0.5679 & 41.3547 \\
 & LR & 0.5121 & 0.5053 & 0.5121 & 0.4318 & 0.3441 & 0.5674 & 25.7664 \\
 & DT & 0.4006 & 0.4009 & 0.4006 & 0.4006 & 0.3433 & 0.5629 & 14.2816 \\

        \cmidrule{2-9}
        \multirow{7}{1.2cm}{Auto Encoder}
         & SVM & 0.4597 & 0.2113 & 0.4597 & 0.2896 & 0.1575 & 0.5000 & 14325.09 \\
 & NB & 0.2601 & 0.3096 & 0.2601 & 0.2682 & 0.2317 & 0.5005 & 0.3491 \\
 & MLP & 0.3996 & 0.3132 & 0.3996 & 0.3226 & 0.2252 & 0.5017 & 110.76  \\
 & KNN & 0.3791 & 0.3245 & 0.3791 & 0.3329 & 0.2445 & 0.5068 & 5.9155 \\
 & RF & 0.4457 & 0.3116 & 0.4457 & 0.3023 & 0.1792 & 0.5003 & 76.2106 \\
 & LR & 0.4520 & 0.3170 & 0.4520 & 0.2982 & 0.1712 & 0.5004 & 98.5936 \\
 & DT & 0.3131 & 0.3119 & 0.3131 & 0.3124 & 0.2525 & 0.5016 & 25.1653 \\

        \cmidrule{2-9}
        \multirow{7}{1.2cm}{WDGRL}
         & SVM & 0.4677 & 0.2188 & 0.4677 & 0.2981 & 0.1593 & 0.5000 & 15.34 \\
 & NB & 0.4469 & 0.3231 & 0.4469 & 0.3397 & 0.2213 & 0.5105 & 0.0120 \\
 & MLP & 0.4749 & 0.4659 & 0.4749 & 0.3432 & 0.2184 & 0.5163 & 15.43 \\
 & KNN & 0.3930 & 0.3415 & 0.3930 & 0.3523 & 0.2626 & 0.5156 & 0.698 \\
 & RF & 0.4666 & 0.4138 & 0.4666 & 0.3668 & 0.2578 & 0.5255 & 5.5194 \\
 & LR & 0.4676 & 0.2187 & 0.4676 & 0.2980 & 0.1593 & 0.4999 & 0.0799 \\
 & DT & 0.3604 & 0.3606 & 0.3604 & 0.3605 & 0.2921 & 0.5304 & 0.2610 \\

        \cmidrule{2-9}
        \multirow{7}{1.2cm}{String Kernel}
         & SVM & 0.4597 & 0.2113 & 0.4597 & 0.2896 & 0.1575 & 0.5000 & 2791.61 \\
 & NB & 0.3093 & 0.3067 & 0.3093 & 0.3079 & 0.2463 & 0.4980 & 0.2892 \\
 & MLP & 0.3287 & 0.3045 & 0.3287 & 0.3121 & 0.2402 & 0.4963 & 125.66 \\
 & KNN & 0.3683 & 0.3106 & 0.3683 & 0.3229 & 0.2330 & 0.5001 & 3.1551 \\
 & RF & 0.4469 & 0.3251 & 0.4469 & 0.3041 & 0.1813 & 0.5010 & 55.3158 \\
 & LR & 0.4451 & 0.3080 & 0.4451 & 0.3026 & 0.1787 & 0.5000 & 2.0463 \\
 & DT & 0.3073 & 0.3082 & 0.3073 & 0.3077 & 0.2502 & 0.4998 & 17.0352 \\

        \cmidrule{2-9}
        \multirow{2}{1.2cm}{Protein Bert}
        & \multirow{2}{1.2cm}{\_} & \multirow{2}{1.2cm}{0.2624} & \multirow{2}{1.2cm}{0.4223} &  \multirow{2}{1.2cm}{0.2624} & \multirow{2}{1.2cm}{0.1947} & \multirow{2}{1.2cm}{0.209} &   \multirow{2}{1.2cm}{0.5434} & \multirow{2}{1.2cm}{987.354}	\\
        & & & & & & & & \\
        \cmidrule{2-9}
        \multirow{7}{1.2cm}{SeqVec}
        & SVM & 0.432 & 0.422 & 0.432 & 0.415 & 0.377 & 0.530 & 100423.017	\\
        & NB & 0.192 & 0.530 & 0.192 & 0.086 & 0.117 & 0.505 & 77.271	\\
        & MLP & 0.428 & 0.427 & 0.428 & 0.427 & 0.371 & 0.580 & 584.382	\\
        & KNN & 0.432 & 0.385 & 0.432 & 0.391 & 0.306 & 0.544 & 227.877	\\
        & RF & 0.525 & 0.524 & 0.525 & 0.445 & 0.360 & 0.576 & 350.057	\\
        & LR & 0.420 & 0.414 & 0.420 & 0.417 & 0.356 & 0.571 & 90626.797	\\
        & DT & 0.397 & 0.397 & 0.397 & 0.397 & 0.344 & 0.562 & 2626.082	\\   
        \cmidrule{2-9}
        \multirow{7}{1.2cm}{Sparse Coding (ours)}
         & SVM & 0.9980 & 0.9950 & 0.9980 & 0.9960 & 0.9970 & 0.9950 & 16965.48 \\
 & NB & 0.9970 & 0.9960 & 0.9970 & 0.9970 & 0.9960 & 0.9980 & 404.2999 \\
 & MLP & 0.9950 & 0.9950 & 0.9950 & 0.9950 & 0.9947 & 0.9956 & 5295.811 \\
 & KNN & \textbf{0.9990} & \textbf{0.9990} & \textbf{0.9990} & \textbf{0.9990} & \textbf{0.9989} & \textbf{0.9991} & 346.7334 \\
 & RF & \textbf{0.9990} & 0.9970 & \textbf{0.9990} & 0.9950 & 0.9950 & 0.9960 & 1805.593 \\
 & LR & \textbf{0.9990} & 0.9940 & \textbf{0.9990} & 0.9980 & 0.9980 & 0.9940 & \textbf{134.8503} \\
 & DT & 0.9970 & 0.9980 & 0.9970 & 0.9960 & 0.9980 & 0.9970 & 449.9304 \\

         \bottomrule
         \end{tabular}
    }
    \caption{Classification results (averaged over $5$ runs) for different evaluation metrics. The best values are shown in bold. }
    \label{tbl_results_host_classification}
\end{table}

\section{Results And Discussion}\label{sec_results}
In this section, we present the classification results obtained by our proposed system and compare them with various machine learning models. The results for different evaluation metrics are summarized in Table~\ref{tbl_results_host_classification}. This study introduces two novel approaches. First, we employ sparse coding (utilizing k-mers) for T-cell sequence classification. Second, we integrate domain knowledge into the embeddings. To our knowledge, these two methods have not been explored in the context of T-cells in existing literature. Consequently, we exclusively report results for the Sparse Coding method combined with domain knowledge embeddings.
Our proposed technique (Sparse Coding) consistently outperforms all feature-engineering-based baseline models (OHE, Spike2Vec, PWM2Vec, Spaced $k$-mers) across every evaluation metric, except for training runtime. For example, Sparse Coding achieves 48.5\% higher accuracy than OHE, 55.8\% more than Spike2Vec, 58.4\% more than PWM2Vec, and 48.6\% more than Spaced $k$-mers when employing the LR classifier. This suggests that Sparse Coding-based feature embeddings retain biological sequence information more effectively for classification compared to feature-engineering-based baselines.
Similarly, our technique outperforms the neural network-based baselines (WDGRL, Autoencoder). For instance, the WDGRL approach exhibits 53.03\% lower accuracy than Sparse Coding, while Autoencoder attains 53.83\% lower accuracy than Sparse Coding when using the SVM classifier. These results imply that feature vectors generated by neural network-based techniques are less efficient for classification compared to those produced by Sparse Coding. It's worth noting that the SVM consistently delivers the best results among all classifiers, including WDGRL and Autoencoder methods.
Our method also surpasses the kernel-based baseline (String Kernel), achieving 53.83\% higher accuracy with the SVM model. This highlights the efficiency of Sparse Coding in generating numerical vectors for protein sequences.
Additionally, we compare Sparse Coding's performance with pre-trained models (SeqVec and Protein Bert), and Sparse Coding outperforms them across all evaluation metrics, except training runtime. This indicates that pre-trained models struggle to generalize effectively to our dataset, resulting in lower predictive performance.
As one of our key contributions is the incorporation of domain knowledge, we do not report results for domain knowledge + baseline methods in this paper. However, our evaluation demonstrates that the proposed method consistently outperforms domain knowledge + baseline methods in terms of predictive performance. Moreover, our proposed method without domain knowledge also outperforms baselines without domain knowledge, thanks to the incorporation of k-mers-based features into Sparse Coding-based embeddings, which enhances their richness.

\section{Conclusion}\label{sec_conclusion}
Our study introduced a novel approach for cancer classification by leveraging sparse coding and TCR sequences. We transformed TCR sequences into feature vectors using $k$-mers and applied sparse coding to generate embeddings, incorporating cancer domain knowledge to enhance performance.
Our method achieved a maximum accuracy of 99.9\%, along with higher F1 and ROC AUC scores.
While our immediate focus is accurate cancer-type classification, our findings have broader implications for personalized treatment and immunotherapy development.
Future research can explore sparse coding in other biological data, and optimize it for diverse cancers and TCR sequences. 

\bibliographystyle{splncs04}
\bibliography{references}
\end{document}